\documentclass{article}

\usepackage[preprint]{neurips_2026}

\usepackage[utf8]{inputenc} 
\usepackage[T1]{fontenc}    
\usepackage{hyperref}       
\usepackage{url}            
\usepackage{booktabs}       
\usepackage{amsfonts}       
\usepackage{nicefrac}       
\usepackage{microtype}      
\usepackage{xcolor}         
\usepackage{amsmath}
\usepackage{multirow}
\usepackage{graphicx} 
\usepackage{fvextra}
\usepackage{listings}
\usepackage{subcaption}
\usepackage{threeparttable}

\title{Genotype-Conditioned Molecular Generation via Evidence-Grounded Multi-Objective Latent Perturbation in Diffusion Models}

\author{%
  Brenda Nogueira
  \\
   Department of Computer Science and Engineering\\
  University of Notre Dame\\
  Notre Dame, IN, 46556 \\
  \texttt{bcruznog@nd.edu} \\
   \And
  Gisela A. González-Montiel
  \\
  Department of Chemistry and Biochemistry\\
  University of Notre Dame\\
  Notre Dame, IN, 46556 \\
  \texttt{ggonzal6@nd.edu} \\
   \AND
  Nitesh V. Chawla
  \\
   Department of Computer Science and Engineering\\
  University of Notre Dame\\
  Notre Dame, IN, 46556 \\
  \texttt{nchawla@nd.edu} \\
   \And
 Nuno Moniz
  \\
   Lucy Family Institute for Data \& Society\\
  University of Notre Dame\\
  Notre Dame, IN, 46556 \\
  \texttt{nunomoniz@nd.edu} \\
}

\begin{document}

\maketitle

\begin{abstract}
 Developing effective anticancer therapeutics remains challenging due to tumor heterogeneity and the absence of well-defined molecular targets across cancer subtypes. Generative models conditioned on cancer genotypes offer a promising avenue for personalized drug discovery, yet existing approaches lack explicit optimization for simultaneous sensitivity, synthesizability, and mechanistic binding plausibility. We present a latent-space optimization approach for a pretrained genotype-to-drug diffusion model, introducing a learnable perturbation over the molecular latent space optimized via gradient ascent to maximize a composite reward combining predicted drug sensitivity (AUC), drug-likeness (QED), and synthetic accessibility (SAS). Critically, biological realism is enforced by grounding both reward design and evaluation in experimentally-derived cancer cell line data and validated pharmacologic signals, anchoring candidate generation in real-world clinical evidence. Mechanistic consistency plausibility is further assessed by a multi-agent LLM pipeline grounded in the diffusion model's attention mechanism. Experiments across 15 cancer cell lines from three held-out evaluation sets demonstrate consistent and noticeable improvements over competing baselines in sensitivity, drug-likeness, synthesizability, and chemical validity.
\end{abstract}

\section{Introduction}

The design of novel anticancer therapeutics remains a central challenge in modern medicine, largely due to tumor heterogeneity and the limited availability of well-defined molecular targets across cancer subtypes~\cite{kim2025genotype}. Recent advances in deep generative models have enabled the exploration of vast chemical spaces, offering a promising paradigm for automated molecular design and personalized drug discovery~\cite{gomez2018automatic,sanchez2018inverse}. In particular, generative approaches conditioned on biological context, such as gene expression or genomic alterations, have shown potential in producing hit-like candidate compounds with target biological traits.

AI-driven drug discovery methods can be broadly categorized into target-based and phenotype-based approaches. The former focuses on generating molecules that bind to predefined proteins, often leveraging docking simulations or reinforcement learning (RL) with binding affinity predictors~\cite{gupta2018generative,renz2019failure,song2023dnmg,chen2025uncertainty,atance2022novo,munson2024novo,bae2023logics}. These approaches rely heavily on prior knowledge of targets, which is often incomplete or unavailable in complex cases such as cancer. In contrast, phenotype-based approaches aim to generate compounds that induce desired cellular responses, bypassing the need for explicit target specification~\cite{das2023gex2sgen,wang2024gldm,liu2024transgem,mendez2020novo}. While promising, these methods typically depend on gene expression data, which can be noisy, biased, and difficult to obtain in clinical settings~\cite{kim2025genotype}.

Studies have explored anti-cancer compound generation conditioned on cancer characteristics using conditional variational autoencoders~\cite{joo2020generative} and RL–based generative models~\cite{park2021molecular}. Recent genotype-conditioned models address some of these issues by using genomic alteration profiles as conditioning signals, improving clinical relevance and interpretability~\cite{kim2025genotype}. Regardless, a critical gap remains: most generative models focus on sampling from learned distributions rather than explicitly optimizing molecules for multiple clinically relevant objectives. For example, effective drug candidates must simultaneously satisfy several criteria, including high sensitivity (efficacy), drug-likeness (QED), synthetic accessibility (SAS), and mechanistic plausibility of target engagement (i.e., binding interactions). Existing approaches often address these objectives through post hoc filtering or weakly integrated reward signals, leading to suboptimal exploration of the molecular space~\cite{born2021paccmannrl,chen2025uncertainty,liu2024multi}.

In this work, we propose a latent-space optimization approach for genotype-conditioned molecular generation that extends pretrained diffusion models with multi-objective optimization and mechanistic reasoning. Rather than relying solely on generative sampling, we introduce a learnable perturbation over the molecular latent space, optimized via gradient-based updates to maximize a composite reward encompassing predicted drug response, physicochemical properties, and mechanistic binding signals. To incorporate biological plausibility, we further extend our approach with a large language model (LLM)-based multi-agent pipeline that evaluates candidate molecules based on literature-grounded plausibility with model-identified targets, grounded in the diffusion model's internal attention mechanisms and real-world feedback from published academic/clinical evidence. The code is available at~\hyperlink{https://anonymous.4open.science/r/GenDiff_Evidence_Opt-1E5B}{https://anonymous.4open.science/r/GenDiff\_Evidence\_Opt-1E5B}. 

Our contributions are summarized as follows:

\begin{itemize}

\item \textbf{Multi-objective latent perturbation with online surrogate critics.} 
We propose a unified perturbation method for frozen genotype-conditioned diffusion models that, unlike related work~\cite{winter2019efficient,liu2024multi}, which requires either optimizing a single differentiable objective or retraining the generative model, jointly optimizes multiple reward axes: predicted sensitivity, drug-likeness, and synthetic plausibility. We address this by training online property surrogate networks in real time on decoded outputs and extending the pattern to sparse LLM feedback via an analogous LLM surrogate critic, enabling dense gradient flow at every optimization step without backbone retraining. Results demonstrate a remarkable improvement over state-of-the-art methods.

\item \textbf{Multi-agent pipeline for mechanistic plausibility scoring.} We propose an agent-based pipeline that 1) extracts per-gene importance scores directly from the genotype conditioning-encoder's internal transformer attention weights, identifying cell-line-specific target genes from NeST adjacency (Biology Agent), then 2) leverages an LLM-based agent to produce a structured non-covalent interaction (NCI) report grounded in retrieved docking literature (Chemistry Agent), then 3) integrating this report with molecular descriptors and biological context using another LLM-based agent, which will produce a scalar mechanistic plausibility score $r_{\text{llm}} \in [0,1]$. This allows the injection of real-world (and verifiable) biomedical knowledge regarding chemical reasoning as a differentiable surrogate approximation of sparse LLM feedback within an optimization loop. Results demonstrate further improvement in key metrics, particularly drug-likeness and synthetic plausibility metrics, as well as on target-engagement generation.

\end{itemize}

\section{Related Work}
\paragraph{Diffusion Models and Multi-Objective Optimization.}

Diffusion models have recently achieved state-of-the-art performance in molecular generation, with geometry-aware and flow-based extensions such as GeoLDM~\cite{xu2023geometric} and PropMolFlow~\cite{zeng2026propmolflow} improving structural validity through SE(3)-equivariant 3D representations. Reinforcement learning (RL) has been widely adopted to guide generation toward drug-relevant objectives, as demonstrated by DrugGen~\cite{sheikholeslami2025druggen}, uncertainty-aware RL-diffusion frameworks~\cite{chen2025uncertainty}, and transformer-based generators such as MolGPT~\cite{bagal2021molgpt} and Taiga~\cite{mazuz2023molecule}. However, RL-based approaches often suffer from reward sparsity, training instability, and overfitting to imperfect predictors. Continuous latent space optimization offers a complementary strategy, with gradient-based and particle swarm methods enabling efficient chemical space navigation~\cite{winter2019efficient} and constrained frameworks enforcing validity during search~\cite{liu2024multi}, yet these approaches typically lack integration of biological context and mechanistic binding rationale into the optimization objective.


\paragraph{Genotype- and Phenotype-Conditioned Molecular Generation.}
Prior work has explored phenotype-conditioned generation using gene expression profiles, with approaches such as PaccMannRL~\cite{born2021paccmannrl} combining transcriptomic inputs with RL to bias generation toward cancer-specific compounds, and subsequent methods extending this to diffusion and latent generative frameworks~\cite{das2023gex2sgen,wang2024gldm,liu2024transgem}. However, gene expression data is noisy, variable, and often impractical in clinical settings. Genotype-conditioned models offer a more clinically grounded alternative: G2D-Diff~\cite{kim2025genotype} conditions diffusion-based generation directly on genomic alteration profiles, producing diverse, drug-like candidates tailored to specific cancer cell lines with interpretable attention mechanisms. Despite these strengths, G2D-Diff optimizes solely for genotype-conditioned sensitivity and lacks multi-objective control over drug-likeness properties such as QED and synthetic accessibility, motivating the optimization approach proposed in this work.

\paragraph{LLMs and Agent-Based Reasoning for Drug Discovery}
Large language models (LLMs) are increasingly applied to drug discovery for molecular representation, reaction prediction, and hypothesis generation~\cite{birhane2023science,zhang2025scientific}, with models such as DrugLLM~\cite{liu2024drugllm} demonstrating few-shot molecular design through learned transformation patterns. Agent-based frameworks, where specialized models collaborate across generation, evaluation, and reasoning roles, have emerged as a promising paradigm for complex scientific tasks~\cite{brown2020language,radford2018improving}, with examples such as scDrugMap integrating multiple large models for multi-modal drug response prediction~\cite{goldstein2024particle}. However, the use of LLMs as mechanistic evaluators within molecular optimization loops remains underexplored — most existing approaches either rely purely on learned predictors or treat LLMs as passive knowledge sources rather than active, biologically-grounded scoring agents. 


\section{Methodology}
\label{sec:methodology}

\begin{figure}[!h]
    \centering
    \includegraphics[width=1.0\linewidth]{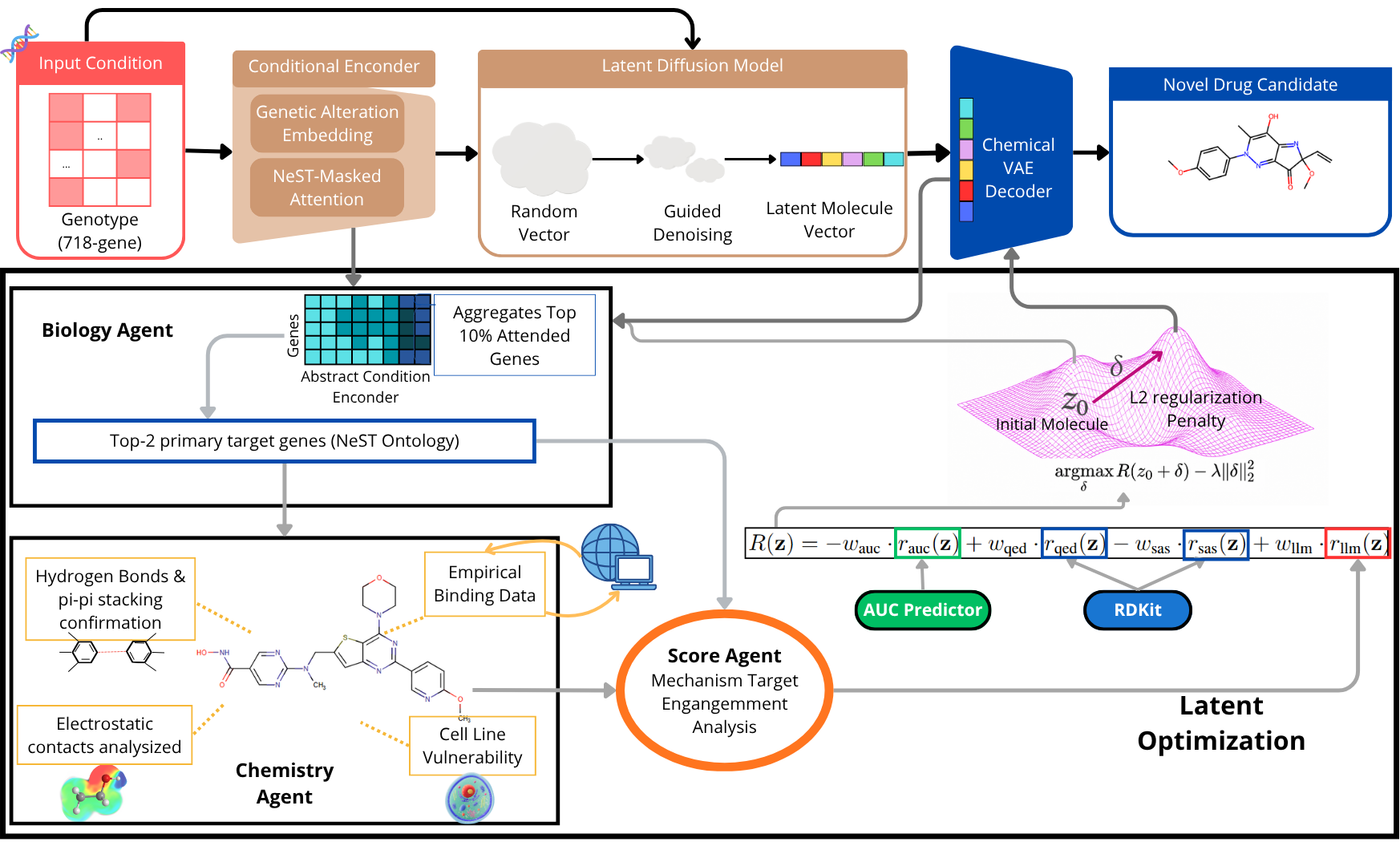}
    \caption{\textbf{Latent Optimization Architecture}: Using a pretrained genotype-conditioning diffusion model, our candidate molecules are passed through a mechanism-scoring pipeline, and summed with drug viability and sensitivity scores, and optimized in the latent space.} 
    \label{fig:architecture}
\end{figure}

We propose a cell-line-conditioned latent space search method for generating molecular candidates that simultaneously optimize three clinically relevant axes: \emph{sensitivity} (predicted drug response as AUC), \emph{drug viability} (drug-likeness as QED and synthesizability as SAS), and \emph{mechanistic target engagement}. The overall approach adopts a multi-agent architecture: a \textbf{BiologyAgent} leverages a conditional diffusion model~\cite{kim2025genotype} to produce all quantitative biological and cheminformatic signals from its internal representations. These signals are then consumed by two LLM-based agents, a \textbf{ChemistryAgent} and a \textbf{ScoreAgent}, that reason over mechanistic plausibility and produce a final binding-effectiveness score. This score is fed back into a latent-space gradient ascent loop, making the overall procedure analogous to a reinforcement learning setting operating over a continuous molecular manifold. Figure~\ref{fig:architecture} summarizes the full pipeline.

\subsection{Optimization Targets}

\paragraph{Sensitivity (AUC).} To assess the sensitivity of a generated candidate, we use a pretrained G2D-Pred~\cite{kim2025genotype}, a genotype-based drug response predictor trained on top of G2D-Diff's condition encoder architecture. G2D-Pred takes the cell-line genotype and a compound latent vector as input and predicts the AUC value of the pair. We use an ensemble predictor of $K=5$ G2D-Pred pre-trained networks with different random seeds; the ensemble mean provides a robust sensitivity estimate:
\begin{equation}
    r_{auc}(\mathbf{z}) = \frac{1}{K}\sum_{k=1}^{K} f_{\theta_k}(\mathbf{z},\, \mathbf{c}),
    \label{eq:auc}
\end{equation}
\noindent with $f_{\theta_k}$ being the k predictor that takes z (latent vector) and c as the genotype condition. In the methodology, lower AUC corresponds to higher sensitivity~\cite{kim2025genotype}.

\paragraph{Drug viability (QED, SAS).} From the decoded SMILES, we use the RDKit package~\cite{landrum2022rdkit} to compute the Quantitative Estimate of Drug-likeness (QED)~\cite{bickerton2012quantifying} and the Synthetic Accessibility Score (SAS)~\cite{ertl2009estimation}, together with a full molecular descriptor profile (MW, LogP, HBD, HBA, TPSA, rotatable bonds, aromatic ring count). The QED and SAS are normalized into rewards $r_{\text{qed}} \in [0,1]$ and $r_{\text{sas}} \in [0,1]$. Note that G2D-Diff itself is trained to generate drug-like molecules and has been shown to produce distributions of QED and SAS comparable to those of reference compound sets~\cite{brown2019guacamol}; our viability rewards further refine this within the optimization loop.

\subsection{Molecule Generator}
For generating our candidates, we use the frozen G2D-Diff~\cite{kim2025genotype}. We selected this model as our backbone because it includes a diffusion process and genotype-conditioning, which are ideal for our pathway enrichment analysis. The model takes as input the binary genotype profile of a cancer cell line. This includes encoding mutations (MUT), copy-number amplifications (CNA), and deletions (CND) across $G$ clinically relevant genes. The encoding process involves two main modules: genetic alteration embedding and a three-layer transformer encoder $\mathcal{E}_\phi$~\cite{vaswani2017attention}. The first transformer layer restricts attention to genes within the same biological subsystem using the NeST ontology~\cite{zheng2021interpretation}, capturing how mutations propagate through interacting pathways. The remaining two layers allow unrestricted global gene-level attention. A response-class token, specifying the desired sensitivity level, is appended to the sequence and acts as a condition vector $\mathbf{c}$ that guides molecule generation~\cite{kim2025genotype}.

The generative backbone is a denoising diffusion probabilistic model (DDPM)~\cite{ho2020denoising}, conditioned on $\mathbf{c}$ via an adaptive instance normalization-based injection module applied in six denoising layers. The pretrained chemical VAE decodes the generated latent vectors to SMILES strings, with a three-layer LSTM encoder-decoder and a 128-dimensional latent space.

\subsection{Agents: Drug-Effectiveness Scoring}\label{sec:agents_method}
\label{sec:llm_agents}


\paragraph{BiologyAgent.} Our BiologyAgent uses the genotype conditional encoder as the source for the biological signals to derive attention-grounded genes. The system extracts internal attention weights directly from the G2D-Diff transformer-based condition encoder $\mathcal{E}_\phi$ to identify true cellular targets. Specifically, we compute per-gene importance scores by averaging the attention from the response-class token to each gene across all heads and layers. Genes are then ranked by their importance scores, and those in the top 10\% that also exceed a uniform attention baseline ($1/(G+1)$, where $G$ is the number of genes) are selected as the \emph{top-attended gene set} (Appendix~\ref{app:attention}). From this set, we further select the top-$k$ genes (with $k=2$ in our experiments) as the primary targets used for downstream reasoning. The raw genotype vectors are parsed to produce a human-readable context comprising the top mutated, amplified, and deleted genes per cell line (from MUT, CNA, and CND, respectively), provided to the LLM agents as supplementary evidence.

\paragraph{ChemistryAgent.} The ChemistryAgent receives the full QED, SAS, molecular descriptors, attention-grounded genes, and gene context and is tasked with generating a \emph{non-covalent interaction (NCI) analysis} report: reasoning over if and how the candidate molecule engages the model-identified target genes. The agent operates in two modes depending on availability. It retrieves literature from PubMed and Semantic Scholar (we do not consider supplementary information of the papers) on AlphaFold structure predictions and molecular docking studies specifically for the attention-grounded top genes, anchoring its NCI analysis in empirical binding data. The agent produces a structured NCI report describing relevant interaction types---hydrogen bonds, $\pi$-$\pi$ stacking, halogen bonds, electrostatic contacts, and hydrophobic interactions---between the candidate and the model-identified targets. Critically, the ChemistryAgent produces a mechanistically grounded NCI evidence report that the next agent can reason over. See the full prompt strategy at GitHub. 

\paragraph{ScoreAgent.} The ScoreAgent integrates all available evidence into a final scalar binding-effectiveness score $r_{\text{llm}} \in [0,1]$. Its inputs are: 1) Molecular profile: QED, SAS, the full molecular descriptors, and predicted AUC; 2) The ChemistryAgent's NCI analysis report, based on surveying the literature; 3) All biology agent outputs: the attention-grounded enriched top-attended genes. The agent outputs structured JSON (example in Appendix~\ref{app:output}) comprising the final score, a confidence estimate, a mechanistic summary sentence, key factors supporting the score, and flagged concerns. 

\subsection{Latent Space Optimization}
\label{sec:latent_opt}

Given a target cell line with condition embedding $\mathbf{c}$, we sample an initial molecule $\mathbf{z}_0 \sim p_\phi(\mathbf{z}|\mathbf{c})$ from the frozen G2D-Diff model: a cell line's genotype-conditioned drug-like candidate. We then introduce a learnable perturbation $\boldsymbol{\delta}$, a vector of equal dimension to $\mathbf{z}_0$ initialized to zero, and optimized by gradient ascent to shift the molecule toward a higher-reward region of the latent space:
\begin{equation}
    \boldsymbol{\delta}^* = \arg\max_{\boldsymbol{\delta}} \; R(\mathbf{z}_0 + \boldsymbol{\delta}) - \lambda\|\boldsymbol{\delta}\|_2^2,
    \label{eq:opt}
\end{equation}
where the $\ell_2$ penalty $\lambda\|\boldsymbol{\delta}\|_2^2$ prevents $\boldsymbol{\delta}$ from growing so large that $\mathbf{z}_0 + \boldsymbol{\delta}$ drifts outside the valid molecular manifold learned by G2D-Diff. The composite reward $R$ balances three objectives:
\begin{equation}
    R(\mathbf{z}) = - w_{\text{auc}} \cdot r_{\text{auc}}(\mathbf{z}) + w_{\text{qed}} \cdot r_{\text{qed}}(\mathbf{z}) - w_{\text{sas}} \cdot r_{\text{sas}}(\mathbf{z}) + w_{\text{llm}} \cdot r_{\text{llm}}(\mathbf{z}).
    \label{eq:reward}
\end{equation}
$r_{\text{qed}} \in [0,1]$ rewards drug-likeness, and $r_{\text{llm}} \in [0,1]$ rewards mechanistic target engagement as assessed by LLM agents. AUC and SAS terms are \emph{subtracted} because $r_{\text{auc}} \in [0,1]$ rewards higher sensitivity (lower AUC means greater drug effect) and $r_{\text{sas}}$ because a higher value means harder to synthesize, i.e., minimizing it improves the reward. We define the weights as $w_{\text{auc}}=1.0, w_{\text{qed}}=0.5, w_{\text{sas}}=0.5, w_{\text{llm}}=1.0$, as we want the reward to reflect in equal parts \emph{sensitivity} (predicted drug response), \emph{drug viability} (drug-likeness and synthesizability), and \emph{mechanistic target engagement}.

\subsubsection{Differentiability and Online Surrogates}

The AUC term $r_{\text{auc}}$ is differentiable through the G2D-Pred ensemble predictors. The drug-viability terms $r_{\text{qed}}$ and $r_{\text{sas}}$ are not directly differentiable, since SMILES decoding via the VAE is discrete. We therefore maintain \emph{online property surrogate networks}, lightweight three-layer MLPs $s_{\text{qed}}(\mathbf{z})$ and $s_{\text{sas}}(\mathbf{z})$ with sigmoid output, trained online via MSE (mean squared error) regression against observed QED/SAS values decoded at each step. Gradients for drug viability flow through these differentiable surrogates. The LLM binding score $r_{\text{llm}}$ is handled analogously: an online LLM surrogate $s_{\text{llm}}(\mathbf{z})$ is updated from the sparse LLM feedback and queried at every gradient step (see Appendix~\ref{app:surrogates}).

\subsubsection{Sparse LLM Feedback and Surrogate Critic}

Because LLM API calls are computationally expensive, the multi-agent pipeline is invoked every $\Delta_{\text{llm}} = 10$ steps on the top-$n$ candidates by predicted AUC. Unscored candidates receive scores imputed via Tanimoto similarity to scored molecules using Morgan fingerprints, with a neutral fallback of $r_{\text{llm}} = 0.5$ when no sufficiently similar scored molecule exists. All scores, direct or imputed, update an online surrogate $s_{\text{llm}}(\mathbf{z})$ that provides a dense, differentiable approximation of $r_{\text{llm}}$ between calls (see Appendix~\ref{app:llm_surrogate}).

\section{Experimental Results}\label{sec:experimental_results}

This section describes our experimental results, first detailing our experimental setup, then focusing on the quality of molecules generated by our overall approach, and finally zooming in on the impact of the proposed multi-agent system. All experiments were conducted on NVIDIA GPU A100 200 GiB RAM, and the LLM used to build the LLM-based agents was Anthropic Claude-sonnet-4-6.

\subsection{Experimental Setup}

\paragraph{Dataset.} We use cell lines in the NCI60 Growth Inhibition Data (accessed July 2022)~\cite{shoemaker2006nci60}, preprocessed at G2D-Diff~\cite{kim2025genotype}. The dataset binary genotype profile of a cancer cell line across $G=718$ clinically relevant genes. Optimization is performed per cell line across 15 cancer cell lines drawn from three held-out evaluation sets (EV1, EV2, EV3): EV1 contains data-rich cell lines seen during training, EV2 contains data-scarce cell lines with known conditions but unseen compound pairs, and EV3 is a fully zero-shot set of unseen cell lines (Appendix~\ref{app:datasets}). We generated 2000 molecular latent vectors. Hyperparameter search for optimization was conducted with all cell lines contained in NCI60 (62), except for those in EV1 and EV2, with a total of 52 cell lines (Appendix~\ref{app:hparams}). 

\paragraph{Baseline comparison.} We used a generative model based on gene expression for hit-like anticancer molecules as baseline (G2D-Diff and PaccMannRL) along with the generation of molecular properties and conditional properties (MolGPT)(Appendix~\ref{app:baselines}), and a known drug-induced growth response dataset (NCI60). We generate and sample around 2000 molecules with each method.


\subsection{Results}

The results below address five questions: (1) does our approach generate higher-quality molecules than state-of-the-art baselines? (2) what is the contribution of each optimization component? (3) how does candidate pool size affect the multi-agent scoring pipeline? (4) does the agent-based scoring demonstrate consistency with expected trends on clinically validated drugs? and (5) are results stable given the stochasticity introduced by LLM-based interactions?

\subsubsection{Quality of Generated Molecules}
\begin{figure}[!h]
    \centering
    \begin{subfigure}{0.4\textwidth}
       
        \includegraphics[width=\linewidth]{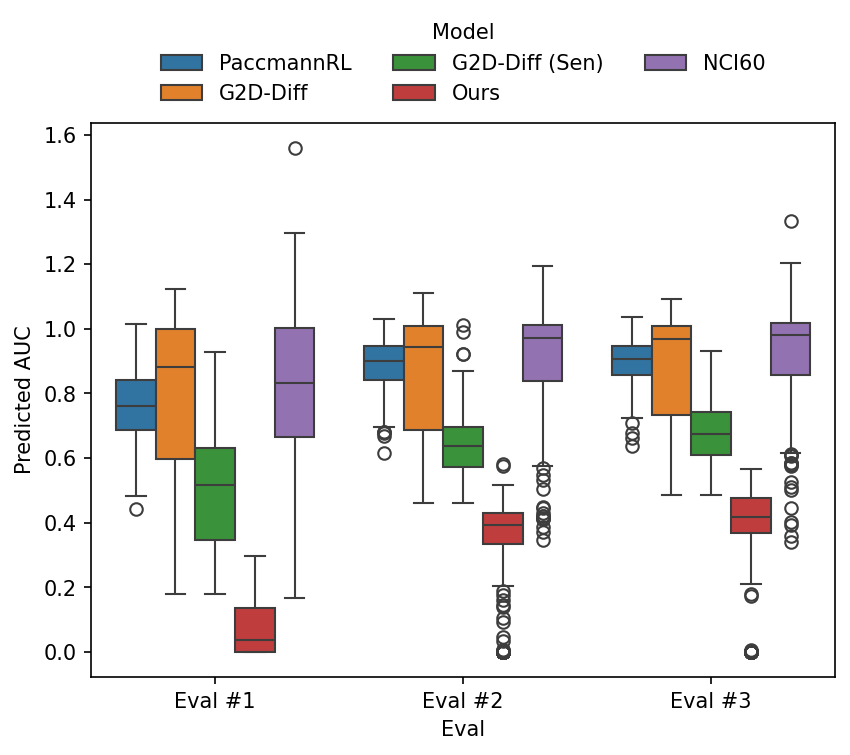}
        
    \end{subfigure}
    \hfill
    \begin{subfigure}{0.58\textwidth}
        \centering
        \includegraphics[width=\linewidth]{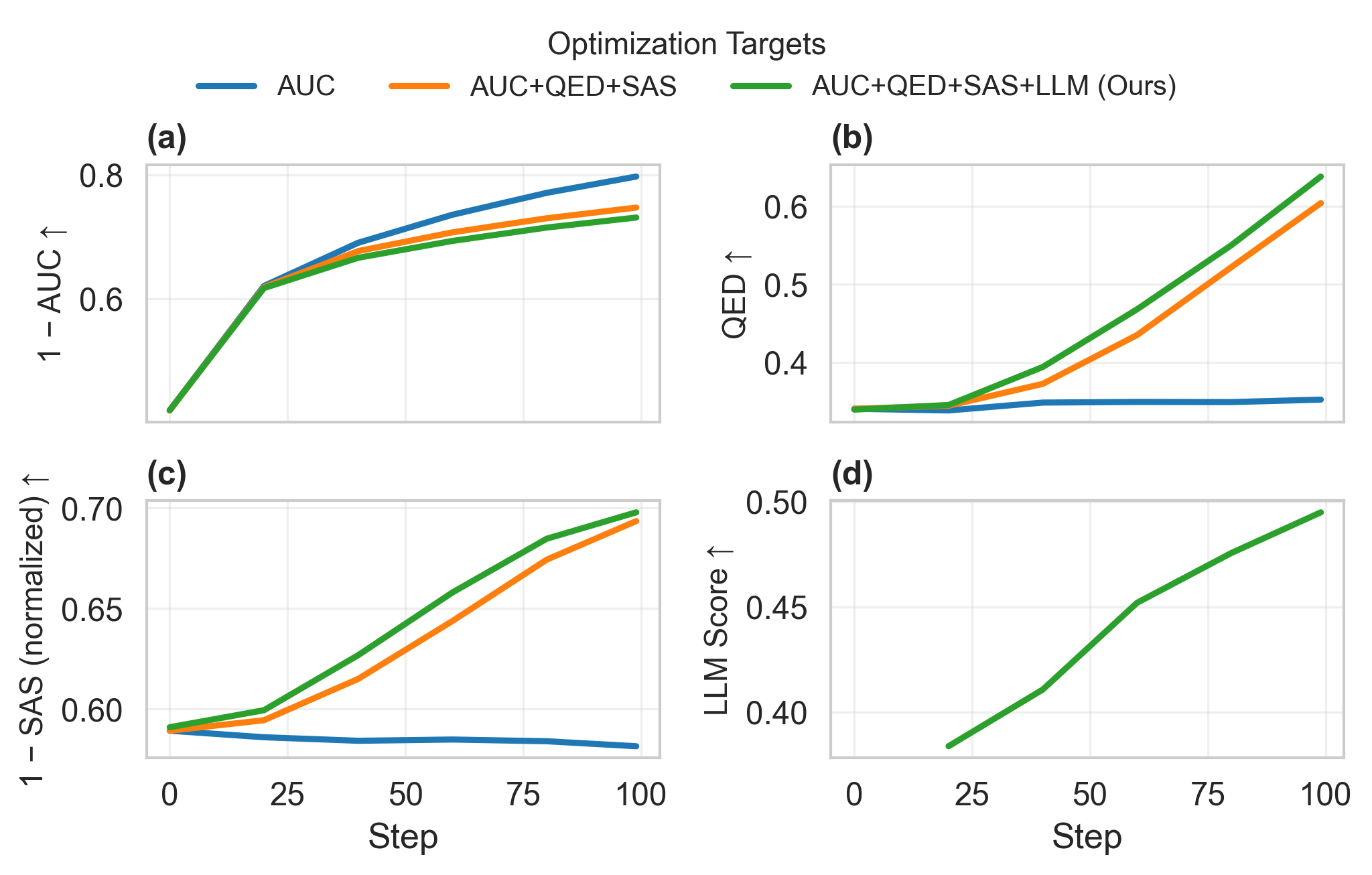}

    \end{subfigure}
    \caption{Latent optimization results.
    (a) Predicted AUC distributions for generated molecules across evaluation sets.  (b) Evolution of AUC, QED, SAS, and agent score across 100 optimization steps for three reward configurations. }
    \label{fig:performance_optimization}
\end{figure}

Table~\ref{tab:performance} and Figure~\ref{fig:performance_optimization} jointly characterize the quality of generated molecules across three evaluation sets for five methods: PaccmannRL, G2D-Diff, and our Agent-optimization latent optimization pipeline (Ours). We also evaluate the results using only the molecules generated with G2D-Diff under very sensitive and sensitive conditions (G2D-Diff (Sen)), and compare the generated molecules with the distribution of real AUC values in the NCI60 dataset. 

Our method achieves the strongest drug-likeness profile across all three evaluation sets, attaining the highest QED scores and the lowest SAS values among the conditional generation baselines. These results suggest the LLM reward signal actively steers optimization toward synthetically feasible, drug-like chemical space. LogP is also the most controlled of the three baselines in all evaluations, suggesting better membrane permeability characteristics and reduced likelihood of promiscuous binders. While MolGen-GPT reports superior QED and SAS, it operates as an unconditional generative model and does not optimize for cell-line-specific sensitivity, making it a reference upper bound for drug-likeness rather than a direct competitor for the conditional generation task. 

Regarding generation quality, our method maintains near-perfect validity while preserving high novelty and competitive uniqueness. The difference in diversity relative to G2D-Diff is consistent with the latent optimization landscape being constrained by the multi-objective reward, which narrows sampling toward a pharmacophore-compatible region of latent space. These results demonstrate that our LLM-augmented pipeline uniquely achieves the combination of high predicted sensitivity, strong drug-likeness, and valid novel chemistry that none of the individual baselines attain simultaneously.

On predicted AUC, our method yields the lowest median predicted AUC across all three evaluation sets. Critically, in the context of drug response, lower AUC values indicate greater predicted sensitivity: our generated molecules are predicted to be substantially more potent against the target cell lines than those produced by competing methods. This sensitivity advantage is most pronounced in Eval \#1, where our method achieves a median predicted AUC of approximately 0.08, versus approximately 0.75 for PaccmannRL and 0.85–0.90 for G2D-Diff. G2D-Diff (Sen) occupies an intermediate position, achieving lower predicted AUCs than G2D-Diff but substantially higher than our method, suggesting that sensitivity conditioning alone, without drug-likeness optimization, generates potent but chemically poor candidates.

\begin{table}[!h]
\caption{Comparison of performance for molecule generation across evaluation sets.}
\label{tab:performance}
\resizebox{\textwidth}{!}{
\begin{tabular}{llcccc|ccc}
\toprule
Eval & Model & Validity ($\uparrow$) & Uniqueness ($\uparrow$) & Novelty ($\uparrow$) & Diversity ($\uparrow$) & SAS ($\downarrow$) & QED ($\uparrow$) & LogP ($\downarrow$)\\
\midrule
 & MolGen-GPT & 0.9849 & \textbf{1.0000} & 0.9653 & 0.8643 & \textbf{2.588} $\pm$ \textbf{0.521} & \textbf{0.829} $\pm$ \textbf{0.076} & \textbf{2.555} $\pm$ \textbf{0.55} \\ \hline
\multirow{3}{*}{Eval \#1} & G2D-Diff & \textbf{1.0000} & 0.9351 & 0.9573 & \textbf{0.9195} & 2.924 $\pm$ 0.861 & 0.454 $\pm$ 0.171 & 3.601 $\pm$ 2.329 \\
 & PaccmannRL & 0.9254 & \textbf{1.0000} & 0.9952 & 0.8351 & 3.702 $\pm$ 0.688 & 0.569 $\pm$ 0.22 & 4.757 $\pm$ 2.684 \\
& \textbf{Ours} & \textbf{1.0000 }& 0.9914 & \textbf{0.9986} & 0.8661 & \textbf{2.873} $\pm$ \textbf{0.613} & \textbf{0.771} $\pm$ \textbf{0.138} & \textbf{3.279} $\pm$ \textbf{1.034 }\\ \hline
\multirow{3}{*}{Eval \#2} & G2D-Diff & \textbf{1.0000} & 0.9663 & 0.9761 & \textbf{0.9137} & 3.264 $\pm$ 1.061 & 0.446 $\pm$ 0.196 & 3.672 $\pm$ 2.305 \\
& PaccmannRL & 0.9299 & \textbf{0.9952} & 0.9968 & 0.8353 & 3.638 $\pm$ 0.704 & 0.587 $\pm$ 0.218 & 4.663 $\pm$ 2.573 \\
 & \textbf{Ours} & 0.9986 & 0.9857 & \textbf{1.0000} & 0.8691 & \textbf{3.076} $\pm$ \textbf{0.733} & \textbf{0.709} $\pm$ \textbf{0.212} & \textbf{3.405} $\pm$ \textbf{1.398} \\ \hline
\multirow{3}{*}{Eval \#3} & G2D-Diff & 0.9983 & 0.9497 & 0.9524 & \textbf{0.9172} & 3.075 $\pm$ 0.963 & 0.46 $\pm$ 0.182 & 3.586 $\pm$ 2.313 \\
 & PaccmannRL & 0.9149 & \textbf{0.9967} & \textbf{0.9967} & 0.8343 & 3.714 $\pm$ 0.717 & 0.584 $\pm$ 0.213 & 4.751 $\pm$ 2.554 \\
 & \textbf{Ours} & \textbf{1.0000} & 0.9900 & 0.9957 & 0.8619 & \textbf{2.854} $\pm$ \textbf{0.641} & \textbf{0.762} $\pm$ \textbf{0.136} & \textbf{3.19} $\pm$ \textbf{1.124} \\
\bottomrule
\end{tabular}}
\end{table}

\subsubsection{Ablation Study}

We conducted ablation studies to assess the contribution of each component in our approach
(Table~\ref{tab:ablation} in Appendix~\ref{app:ablation}). It can be seen that every component of our method has an important impact on our multi-objective optimization, and our method consistently outperforms across key metrics (SAS, QED, LogP, LLM Score). Although uniqueness, novelty, and diversity slightly
decrease, the trade-off is acceptable given the substantial gains in other objectives. Simplified
variants of our method show clear performance improvements, confirming the necessity of the full design.

Figure~\ref{fig:performance_optimization} also presents the evolution of latent optimization reward configurations across 100 optimization steps. It shows that optimizing for AUC alone achieves the highest raw sensitivity score, but at the cost of drug-likeness: QED remains flat and synthetic accessibility degrades slightly throughout optimization. Adding QED and SAS as explicit reward terms substantially improves drug-likeness while only modestly reducing AUC performance, demonstrating the expected Pareto trade-off between potency and drug-likeness. Our full agent-augmented method achieves the best overall profile: it matches or exceeds QED and SAS optimization, while additionally optimizing biological relevance as measured by the LLM scorer.

\subsubsection{Sensitivity to Candidate Molecules Scored}

Because LLM API calls are computationally expensive, the multi-agent scoring pipeline is invoked periodically and applied only to the top-$n$ candidate molecules ranked by predicted AUC at each optimization step. To evaluate the impact of this design choice, we performed latent optimization using different candidate pool sizes ($n \in {5,10,25,50}$) presented in Table~\ref{tab:samples_n} using our 3 evaluation sets (Appendix~\ref{app:datasets}). The results indicate that increasing the number of sampled candidates does not consistently improve optimization performance. While larger candidate pools provide a broader search space, gains beyond 25 samples are marginal or inconsistent across metrics. In particular, the 25-sample configuration achieves the best overall trade-off, obtaining the highest mean LLM score, QED, and reward, while maintaining competitive AUC and validity. This suggests that moderate candidate pool sizes are sufficient to capture the benefits of LLM-guided re-ranking, whereas further increases incur additional computational cost without proportional improvements. These findings support the use of 25 candidates as an effective balance between optimization quality and computational efficiency.
\begin{table}[!h]
    
     \caption{Final performance (step = 100) across different sample sizes (5, 10, 25, 50). We report average values over all cell lines for AUC, QED, SAS (normalized), and LLM score, along with loss, validity, and mean reward.}
    \label{tab:samples_n}
    \centering
    \resizebox{\textwidth}{!}{
   \begin{tabular}{llp{2cm}lllll}
   \toprule 
   Model & Loss $\downarrow$ & Mean LLM Score $\uparrow$ & Mean AUC $\downarrow$ & Mean QED $\uparrow$ & Mean SAS $\downarrow$& Validity $\uparrow$& Mean Reward $\uparrow$\\
   \midrule
   5 & -0.277071 & 0.464271 & 0.268517 & 0.631614 & \textbf{2.954874} & \textbf{0.938571} & 0.312650 \\ 
   10 & -0.266011 & 0.473477 & \textbf{0.267920} & 0.619528 & 2.984952 & 0.920476 & 0.301915 \\
   25 & \textbf{-0.299841} & \textbf{0.494878} & 0.268998 & \textbf{0.638203} & 3.021048 & 0.920000 & \textbf{0.335169} \\
   50 & -0.271069 & 0.462252 & 0.269359 & 0.630753 & 2.976875 & 0.928571 & 0.306834 \\
   \bottomrule
   \end{tabular}}
\end{table}

\subsubsection{Clinically Validated Drugs Evaluation}\label{sec:val_drugs}
We constructed an evaluation suite of 11 compound–cell-line pairs covering four FDA-approved targeted therapies (See Appendix~\ref{app:val_drugs}). The results are presented in Table~\ref{tab:Known Binders}. The Agent-based scoring component shows consistency with expected trends, assigning higher scores to known actives and lower scores to negative or irrelevant compounds. Erlotinib is an exception, as its drug-like kinase scaffold scores well on NCI features despite targeting EGFR rather than BCR-ABL in K562. Our Spearman correlation test demonstrates a significant correlation between the LLM Score and the estimated expected values (correlation=0.784, $p_{value}=4.27e-03$). Notably, these trends are also aligned with NCI similarity assessments, suggesting that the LLM captures chemically and biologically meaningful features that are missed by the AUC estimation pipeline. Together, these results indicate that while AUC predictions using the molecules encoded by the VAE model may be unreliable due to encoder-induced biases (Appendix~\ref{app:encoder_bias}), LLM-based evaluations provide a more robust and interpretable signal of compound relevance. This highlights the importance of incorporating alternative evaluation mechanisms and motivates the development of biologically grounded and relevance-aware scoring strategies.

\begin{table}[!h]
    \centering
    \caption{LLM scores for clinically validated drug binders across cancer cell lines.}
    \label{tab:Known Binders}
        \begin{threeparttable}

    \resizebox{\textwidth}{!}{
\begin{tabular}{llrrrllr}
\toprule
Name & Cell Name & Pred AUC~\tnote{1} & QED & SAS & Expected LLM Score & NCI Similarity & LLM Score \\
\midrule
imatinib  & K562 & 0.555 & 0.389 & 2.332 & high & moderate & 0.380 \\
dasatinib  & K562 & 0.527 & 0.463 & 2.675 & high & moderate & 0.420 \\
alpelisib  & MCF7 & 0.816 & 0.393 & 2.533 & high & moderate & 0.360 \\
vemurafenib  & LOXIMVI & 0.799 & 0.261 & 2.415 & high & moderate & 0.310 \\
wortmannin  & MCF7& 0.845 & 0.555 & 5.634 & medium & low & 0.280 \\
erlotinib  & K562 & 0.771 & 0.526 & 2.506 & medium-low & moderate & 0.580 \\
aspirin  & K562 & 0.792 & 0.550 & 1.580 & low & low & 0.180\\
metformin  & K562 & 1.003 & 0.249 & 3.674 & low & low & 0.280 \\
caffeine  & K562 & 0.830 & 0.538 & 2.298 & low & low & 0.210 \\
PAINS quinone  & MCF7 & 0.827 & 0.575 & 1.926 & low & low & 0.280 \\
hexane  & K562 & 0.785 & 0.463 & 1.209 & very low & low & 0.040 \\
\bottomrule
\end{tabular}}

   \begin{tablenotes}
\footnotesize
\item[1] \parbox{0.8\linewidth}{
Predicted AUC values for known binders are systematically elevated due to the frozen G2D-Diff VAE encoder, which is not optimized on drug–response data and may map compounds to latent regions misaligned with the sensitivity-conditioned subspace, inflating AUC estimates (see Appendix~\ref{app:encoder_bias}).
}
\end{tablenotes}
    
    \end{threeparttable}
\end{table}

\subsubsection{Stability of LLM-Based Scoring}\label{sec:score_analysis}

\begin{figure}[!h]
     \centering
      \begin{subfigure}[b]{0.49\textwidth}
         \centering
         \includegraphics[width=\textwidth]{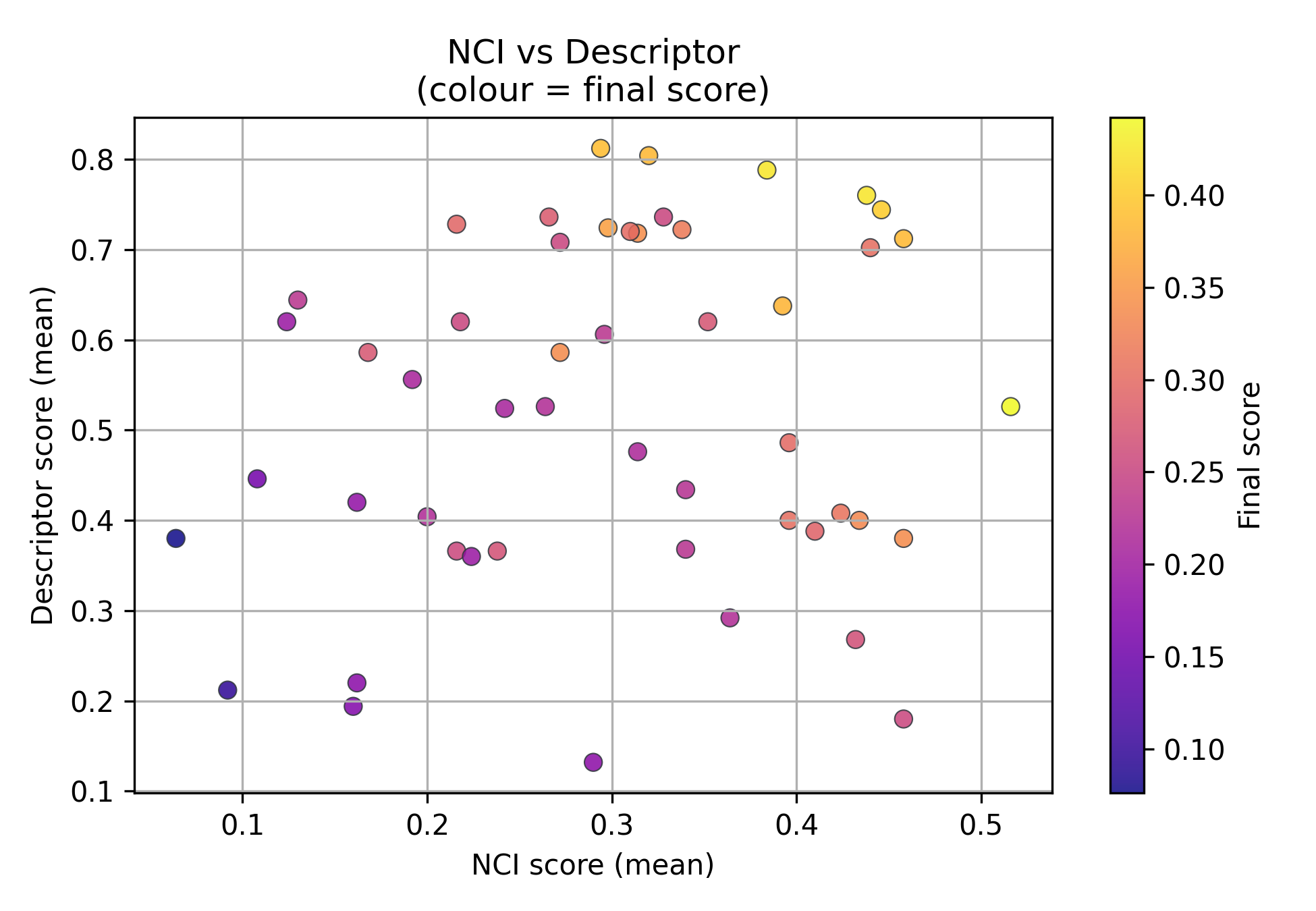}
     \end{subfigure}
     \hfill
     \begin{subfigure}[b]{0.49\textwidth}
         \centering
         \includegraphics[width=\textwidth]{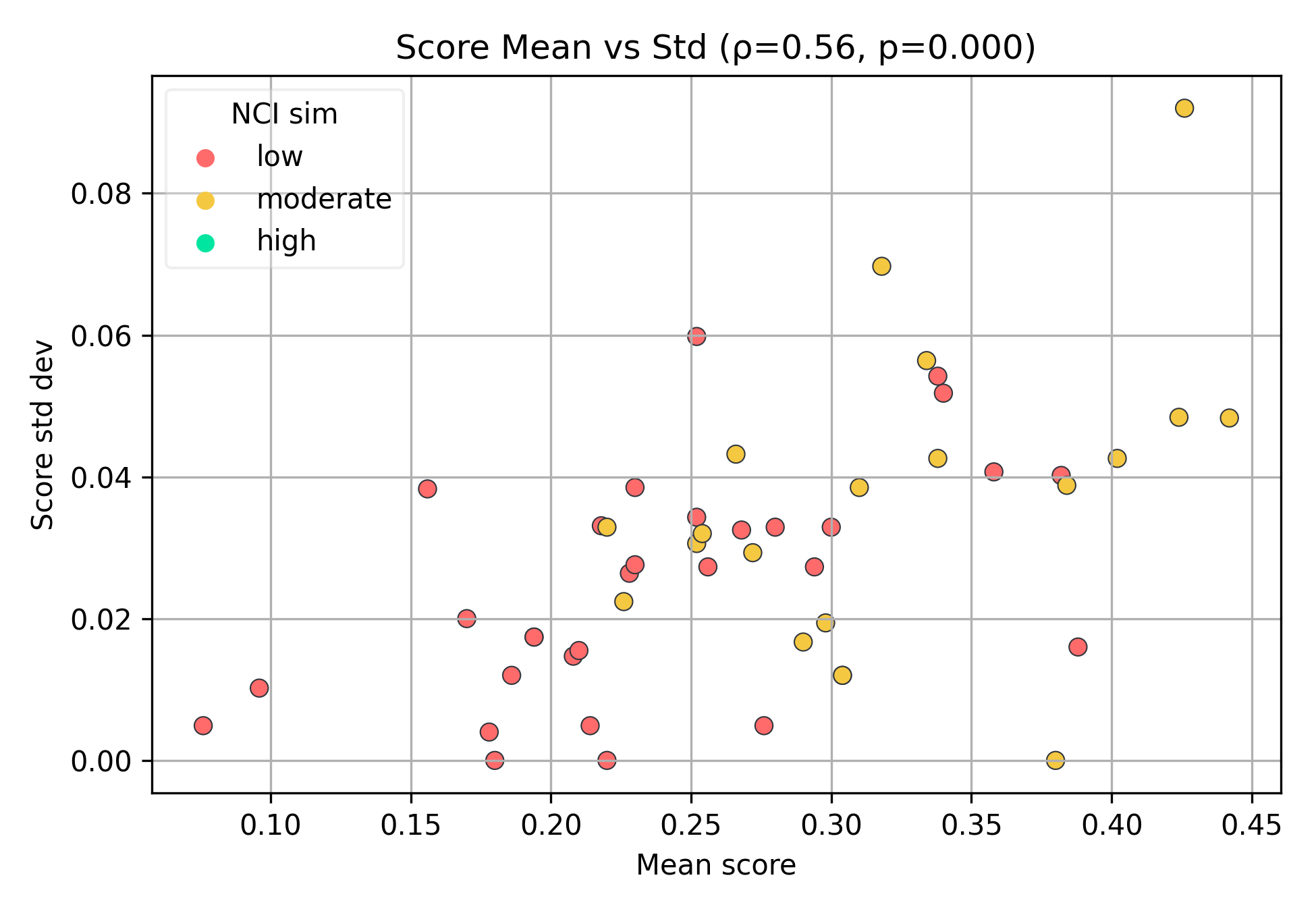}
     \end{subfigure}
     
     \caption{Score relationships. (a) Relationship between mean NCI score and descriptor score. (b) Consistency of scores across repeated evaluations.}
     \label{fig:nci}
\end{figure}

Our multi-agent pipeline was evaluated for stability on 50 randomly sampled cell-line pairs, each scored five times under identical conditions. The mean standard deviation of the final score across repetitions is 0.03, confirming high consistency. Score component analysis reveals a systematic asymmetry (Figure~\ref{fig:nci}): descriptor scores are consistently higher than NCI scores, reflecting that unoptimized diffusion samples are generally drug-like but exhibit weak binding evidence for the model-identified target genes: precisely the gap that latent-space optimization addresses. Notably, molecules with moderate NCI overlap exhibit higher inter-run variability, suggesting that NCI similarity influences both score magnitude and stability, and that improvements in NCI overlap during optimization may increase both score variance and mean score. Recommendation stability is 88\% and target gene pair identification is consistent in 80\% of cases, confirming reliable qualitative outputs.

We also assessed Pearson correlations between molecular properties, predicted AUC, and the final LLM score. QED and SAS show statistically significant correlations ($r=0.356$, $p_{value}=0.011$ and $r=-0.283$, $p_{value}=0.046$ respectively), confirming that the descriptor-based component contributes meaningfully to the overall score, while predicted AUC shows no significant correlation ($r=-0.218$, $p_{value}=0.127$). This indicates that the LLM score captures additional mechanistic and contextual information beyond the sensitivity predictor. Literature grounding is robust: only two gene pairs returned zero supporting papers, with an average of 11.04 papers per pair and 7.08 full-text articles, confirming that the pipeline anchors its reasoning in existing biomedical evidence.

\section{Conclusion}
In this work, we presented a genotype-conditioned molecular generation and multi-objective optimization that combines diffusion-based latent space optimization with an LLM-augmented biological scoring pipeline for cancer cell-line-specific drug design. Building on a pretrained genotype-conditioned generator, we introduce a multi-objective reward that jointly optimizes predicted drug sensitivity (AUC), drug-likeness (QED), synthetic accessibility (SAS), and a biologically grounded agent score. This was achieved by establishing a multi-agent system (Biologist, Chemistry, and Score Agents). We have extensively demonstrated how the proposed approach consistently achieves the lowest predicted AUC, indicating improved cell-line specificity, while also matching or exceeding conditional baselines in QED and SAS, with near-perfect validity and novelty. Latent-space optimization experiments show that incorporating the LLM agent score accelerates convergence toward biologically meaningful chemical space. Clinical validated drugs evaluation further confirms that the full pipeline, combining AUC prediction, RDKit descriptors, and gene-context attention, effectively ranks FDA-approved targeted therapies above decoys and reactive compounds. The positive impact of our work lies in reducing the cost and time required for in-vitro experiments to identify early-stage drug candidates for cancers with complex or poorly understood mutational landscapes. Acknowledging that generative models can be misused to design compounds with unintended biological activity, our approach still requires wet-lab validation, only producing candidates that are evaluated rather than synthesized. 

\paragraph{Limitations and Future Work}
A key limitation in our pipeline is the LLM scorer's dependence on the calibration quality of the upstream AUC predictor. Meaning, when sensitivity estimates are unreliable, biological reasoning cannot fully compensate. A second limitation in our work is that NeST is currently used only at the gene-level adjacency structure, without fully exploiting its pathway ontology for higher-level biological reasoning. Future work will extend the proposed approach to incorporate pathway-level reasoning, cross-pathway interactions, and hierarchical biological structure. Finally, we plan to 
incorporate structure-based validation using AlphaFold-derived docking estimates. 
Importantly, we will focus on wet-lab validation, aiming for real-world validation of our approach.

\begin{ack}
    The project was partially supported by the National Science Foundation under the NSF Center for Computer Assisted Synthesis (C-CAS; grant no. CHE-2202693).
\end{ack}

\bibliographystyle{abbrv}
\bibliography{sample}

\appendix

\section{Attention-Grounded Target Identification}
\label{app:attention}

\subsection{Overview}

To identify which genes G2D-Diff associates with drug sensitivity for a given cell line, we extract and analyse the attention weights produced by the condition encoder $\mathcal{E}_\phi$ during a forward pass. This procedure follows the interpretability methodology described in~\cite{kim2025genotype}. The output feeds directly into the ChemistryAgent prompt as grounded biological evidence, replacing reliance on LLM prior knowledge about target genes.

\subsection{Attention Extraction}

The condition encoder $\mathcal{E}_\phi$ is instantiated in attention-logging mode, creating a weight-sharing copy of the original frozen encoder. A forward pass with the cell-line genotype batch returns a four-tuple:
\[
(\mathbf{F},\; \mathbf{c},\; \mathbf{a}_{\text{last}},\; [\mathbf{A}^{T_{\text{neigh}}},\, \mathbf{A}^{T_{\text{whole}}},\, \mathbf{A}^{T_{\text{reout}}}])
\]
where $\mathbf{F} \in \mathbb{R}^{B \times 719 \times 128}$ is the full token sequence, $\mathbf{c} \in \mathbb{R}^{B \times 128}$ is the condition embedding (CLS token), $\mathbf{a}_{\text{last}}$ is the last-layer CLS-row attention, and $\mathbf{A}^{(\cdot)} \in \mathbb{R}^{B \times H \times (G+1) \times (G+1)}$ are the full attention tensors from each of the three transformer blocks ($G=718$ genes, $+1$ for the CLS token, $H$ heads).

\subsection{Gene Importance Aggregation}

Per-gene importance scores are computed following the paper's procedure:

\begin{enumerate}
    \item For each layer $l \in \{T_{\text{neigh}}, T_{\text{whole}}, T_{\text{reout}}\}$ and each attention head $h$, extract the CLS-to-gene attention row $\mathbf{A}^{(l)}_{[0, 1:G+1]} \in \mathbb{R}^G$ (token 0 is CLS).
    \item Apply a uniform attention threshold: retain only entries exceeding $1/(G+1)$, the value that would result from uniform attention over all tokens.
    \item Average the retained scores across heads $H$ to obtain a per-layer gene score $\mathbf{s}^{(l)} \in \mathbb{R}^G$.
    \item Average across all three layers: $\bar{\mathbf{s}} = \frac{1}{3}\sum_l \mathbf{s}^{(l)}$.
\end{enumerate}

The \emph{top-attended gene set} $\mathcal{G}^*$ is defined as genes in the top 10\% of $\bar{\mathbf{s}}$ that additionally exceed the uniform baseline:
\begin{equation}
    \mathcal{G}^* = \left\{ g \;\middle|\; \bar{s}_g \geq \text{top-}10\%(\bar{\mathbf{s}}) \;\text{ and }\; \bar{s}_g > \frac{1}{G+1} \right\}.
\end{equation}



\subsection{Output}

The attention result is formatted into a structured block injected into the ChemistryAgent prompt, comprising: (i) top-attended genes with their exact attention scores,  and (iii) the priority intersection genes explicitly flagged for docking literature search. 
\section{LLM Agent Output}
~\label{app:output}

\begin{lstlisting}[
basicstyle=\tiny,breaklines=true,breakatwhitespace=true
]
21:04:50 | INFO | Web search ENABLED for ChemistryAgent
21:04:50 | INFO | Cell:      HS578T_BREAST
21:04:50 | INFO | Mutated:   ['DIS3L2', 'PIK3R1', 'SLIT2', 'TP53', 'HRAS', 'ING1', 'PTCH1', 'HNF1A']
21:04:50 | INFO | Amplified: ['PREX2', 'ELOC', 'RNF139', 'RUNX1T1', 'RECQL4', 'C8orf34', 'RAD21']
21:04:50 | INFO | Deleted:   ['SMARCA1', 'AMER1', 'PAK3', 'BTK', 'DMD', 'DKC1', 'PHF6']
21:04:50 | INFO | Loading diffusion model from ./data/model_ckpts/1229_512_adanorm_6layers_2474.ckpt ...
NeST neighbor info is used
Load pretrained cond_encoder ...
NeST neighbor info is used
21:04:56 | INFO | ConditionEncoderWithAttention: created get_att=True encoder copy
21:04:56 | INFO | NeST adjacency loaded for hybrid re-ranking: (718, 718)
21:04:56 | INFO | Loaded 3 NeST gene sets from ./data/NeST_neighbor_adj.npy
21:04:56 | INFO | Attention extracted: 71 top genes, 0 enriched pathways
21:04:56 | INFO |   Top genes: ['TSHR', 'PTCH1', 'PIK3R1', 'RANBP2', 'GABRA6', 'ROS1', 'ABCC3', 'ARAF']
21:04:56 | INFO | 
Scoring: CC(C)(C[C@@](CC1=CC2=C(N1)C=NC=C2)(C(F)(F)F)O)C3=C(C=CC(=C3)F)O
21:04:56 | INFO | Pipeline: ChemistryAgent (NCI analysis, attention_grounded=True) ScoreAgent (final score)
21:04:56 | INFO |   [literature] searching for TSHR...
21:04:57 | INFO |   [literature] retrying with broader query for TSHR...
21:05:09 | INFO |   [literature] TSHR: 4 paper(s), 2 with full text, 2 abstract-only
21:05:09 | INFO |   [literature] searching for PTCH1...
21:05:11 | INFO |   [literature] PTCH1: 2 paper(s), 1 with full text, 1 abstract-only
21:05:11 | INFO |   [ChemistryAgent] literature retrieved, injecting into prompt
{
  "smiles": "CC(C)(C[C@@](CC1=CC2=C(N1)C=NC=C2)(C(F)(F)F)O)C3=C(C=CC(=C3)F)O",
  "pred_auc": 0.3,
  "qed": 0.5518316233119285,
  "sas": 3.3556174981731157,
  "attention_grounded": true,
  "final_score": 0.61,
  "nci_score": 0.55,
  "descriptor_score": 0.7,
  "confidence": 0.58,
  "recommendation": "investigate_further",
  "summary": "The candidate shows strong predicted sensitivity (AUC=0.30) in HS578T breast cancer cells and carries reasonable drug-likeness (QED=0.552, SAS=3.36) with no Lipinski violations. NCI overlap with TSHR and PTCH1 known binders is moderate: the fluorophenol-indole-pyridine scaffold supports pi-pi stacking and HBD interactions, but the bulky CF3-quaternary carbon likely clashes with narrow binding pockets and the absence of a basic amine limits salt-bridge formation seen in high-affinity hits. The PTCH1-mutant Hedgehog dependency in HS578T provides a plausible genotype-matched mechanism, though structural gaps versus vismodegib and known PTCH1 binders temper confidence.",
  "key_factors": [
    "Excellent predicted sensitivity: AUC=0.30 in PTCH1/HRAS/TP53-mutant HS578T",
    "Fluorophenol pharmacophore mimics hydroquinone PTCH1 binders; indole/pyridine support pi stacking",
    "Good synthetic accessibility (SAS=3.36) and QED=0.552 support optimization tractability"
  ],
  "concerns": [
    "Bulky CF3-quaternary carbon likely causes steric clash in TSHR TMD and PTCH1 Neck pocket",
    "Absence of basic amine weakens binding vs. known TSHR hits; moderate LogP may increase off-target hydrophobic binding"
  ],
  "chemistry_report": {
    "target_genes_with_docking_data": [
      "TSHR",
      "PTCH1"
    ],
    "attention_grounded": true,
    "nci_analysis": [
      {
        "gene": "TSHR",
        "attention_score": "0.0765",
        "in_enriched_pathway": false,
        "known_hits": [
          "NCATS-SM4420 (A35)",
          "NCGC00241808",
          "org43553",
          "compound 2"
        ],
        "key_ncis": {
          "pi_pi": "Small-molecule agonists stack with aromatic residues in TMD serpentine binding pocket",
          "hydrogen_bonding": "NH/OH donors interact with polar TMD residues (e.g., D619, N674) in transmembrane cavity",
          "halogen_bonding": "Fluorinated analogues show halogen contacts with backbone carbonyls in TMD",
          "electrostatic": "Protonatable amine groups form salt bridges with acidic TMD residues",
          "hydrophobic": "Aliphatic and aromatic groups pack into hydrophobic TMD pocket lining helices 3-6"
        },
        "candidate_nci_overlap": "Candidate has 3 aromatic rings, 3 HBDs (OH, NH), and 2 fluorines for pi-pi and halogen interactions",
        "candidate_nci_gaps": "Candidate lacks basic amine for salt bridge; high LogP may limit selectivity over hydrophobic TMD pocket"
      },
      {
        "gene": "PTCH1",
        "attention_score": "0.0501",
        "in_enriched_pathway": false,
        "known_hits": [
          "panicein A hydroquinone (PAH)",
          "doxorubicin (efflux substrate)",
          "vismodegib (pathway)"
        ],
        "key_ncis": {
          "pi_pi": "Aromatic hydroquinone ring stacks within PTCH1 Neck pocket hydrophobic cavity",
          "hydrogen_bonding": "Hydroquinone OH donors hydrogen bond with polar Neck pocket residues",
          "halogen_bonding": "Not prominently described for PAH analogues in PTCH1 Neck pocket",
          "electrostatic": "Limited charged interactions; primarily hydrophobic Neck pocket engagement",
          "hydrophobic": "Cylindrical open conformation of ligand fills elongated hydrophobic Neck pocket"
        },
        "candidate_nci_overlap": "Candidate's phenol OH and indole NH provide HBD matching; fluorophenol mimics hydroquinone pharmacophore",
        "candidate_nci_gaps": "Bulky tert-butyl/CF3 group may clash with narrow Neck pocket; candidate lacks cylindrical shape preferred by PTCH1"
      }
    ],
    "overall_nci_similarity": "moderate",
    "structural_rationale": "The candidate's fluorophenol, indole, and pyridine rings support pi-pi stacking and HBD interactions analogous to known TSHR TMD and PTCH1 Neck pocket binders. However, the CF3-quaternary carbon bulk and absence of a basic amine may limit optimal fit in both target cavities.",
    "genotype_exploitation": {
      "mechanism": "pathway_dependency",
      "key_gene": "PTCH1",
      "known_therapy_analogy": "vismodegib",
      "rationale": "PTCH1 mutation in HS578T activates Hedgehog signaling; candidate's hydroquinone-like pharmacophore may inhibit PTCH1 efflux function, complementing TP53 mutation-driven genomic instability and HRAS-driven proliferation in this breast cancer cell line."
    }
  }
}
\end{lstlisting}
\section{Online Property Surrogate Networks}
\label{app:surrogates}

\subsection{Motivation}

Two of the three reward components — QED and SAS — require decoding a latent vector $\mathbf{z}$ into a SMILES string and then applying RDKit computations. This process is non-differentiable, which prevents gradient-based optimisation from directly using these signals. We address this with \emph{online property surrogate networks}: lightweight MLPs that are trained in real time on observed (latent, property) pairs and provide differentiable approximations of QED and SAS throughout the optimisation loop.

\subsection{Architecture}

Each surrogate $s: \mathbb{R}^{d} \to [0,1]$ is a three-layer MLP:
\begin{equation}
    s(\mathbf{z}) = \sigma\bigl(W_3 \,\text{ReLU}(W_2 \,\text{ReLU}(W_1 \mathbf{z} + b_1) + b_2) + b_3\bigr),
\end{equation}
where $d$ is the VAE latent dimension (128), hidden layers have dimensions 128 and 64, and $\sigma$ is the sigmoid activation that maps the output to $[0,1]$. Two separate surrogates are maintained: $s_{\text{qed}}$ and $s_{\text{sas}}$. Note that for $s_{\text{sas}}$, the raw SAS score is normalised by dividing by 10 before being used as the training target, so the surrogate always predicts in $[0,1]$ regardless of which property it approximates.

\subsection{Online Training Procedure}

At each optimisation step $t$:
\begin{enumerate}
    \item Decode the current batch $\{\mathbf{z}_i\}_{i=1}^N$ via the VAE decoder to obtain SMILES strings.
    \item Compute ground-truth RDKit properties: $\text{QED}_i$ and $\text{SAS}_i / 10$.
    \item For invalid SMILES, assign fallback values: QED $= 0$, SAS/10 $= 1$.
    \item Update each surrogate for $n_{\text{inner}}$ gradient steps using MSE loss:
    \begin{equation}
        \mathcal{L}_{\text{qed}} = \frac{1}{N}\sum_{i=1}^N \bigl(s_{\text{qed}}(\mathbf{z}_i^\text{sg}) - \text{QED}_i\bigr)^2,
    \end{equation}
    where $\mathbf{z}_i^\text{sg}$ denotes a stop-gradient copy of $\mathbf{z}_i$, ensuring surrogate updates do not interfere with the main optimisation graph.
\end{enumerate}

\paragraph{Warmup period.} During the first $\tau_{\text{warm}} = 5$ optimisation 
steps, the surrogates have seen too few data points to be reliable. During this 
warmup, only the AUC gradient is used. The full reward at step $t$ is:
\begin{equation}
    R_t(\mathbf{z}) = \begin{cases}
        -w_{\text{auc}} \cdot r{a}(\mathbf{z})
        & t < \tau_{\text{warm}}, \\[6pt]
        -w_{\text{auc}} \cdot r(\mathbf{z})
        + w_{\text{qed}} \cdot s_{\text{qed}}(\mathbf{z})
        - w_{\text{sas}} \cdot s_{\text{sas}}(\mathbf{z})
        & \tau_{\text{warm}} \leq t < \tau_{\text{llm}}, \\[6pt]
        -w_{\text{auc}} \cdot r{a}(\mathbf{z})
        + w_{\text{qed}} \cdot s_{\text{qed}}(\mathbf{z})
        - w_{\text{sas}} \cdot s_{\text{sas}}(\mathbf{z})
        + w_{\text{llm}} \cdot s_{\text{llm}}(\mathbf{z})
        & t \geq \tau_{\text{llm}},
    \end{cases}
\end{equation}
where $\tau_{\text{llm}}$ denotes the first step at which the LLM surrogate 
$s_{\text{llm}}(\mathbf{z})$ has received sufficient updates to be reliable 
(i.e., after the first LLM invocation at step $\Delta_{\text{llm}}$). The 
negative sign on $r{a}(\mathbf{z})$ reflects that AUC is minimized --- 
lower predicted AUC corresponds to greater predicted drug sensitivity --- while 
QED and LLM score are maximized and SAS is minimized (lower SAS indicating 
easier synthesis).

\subsection{LLM Surrogate}
\label{app:llm_surrogate}

An online surrogate $s_{\text{llm}}(\mathbf{z})$ is maintained to convert sparse,  expensive LLM feedback into a dense, differentiable reward signal. It shares the same architecture as the QED and SAS surrogates --- a three-layer MLP with ReLU activations and a sigmoid output --- and maps the latent vector $\mathbf{z}$  directly to a score in $[0, 1]$.

\paragraph{Invocation schedule.}
The multi-agent scoring pipeline is invoked every $\Delta_{\text{llm}} = 10$ optimization steps, applied only to the top-$n$ candidates ranked by predicted AUC at that step, balancing evaluation quality against API cost.

\paragraph{Score imputation.}
Candidates not directly evaluated by the LLM receive imputed scores via Tanimoto similarity to the scored molecules, computed from Morgan fingerprints on the decoded SMILES. Formally, for each unscored candidate $\mathbf{z}_j$:
\begin{equation}
    r_{\text{llm}}(\mathbf{z}_j) =
    \begin{cases}
        \dfrac{\sum_{i \in \mathcal{S}} \text{Tan}(j, i)\, \hat{r}_i}
              {\sum_{i \in \mathcal{S}} \text{Tan}(j, i)}
        & \text{if } \max_{i \in \mathcal{S}}\, \text{Tan}(j,i) > \tau_{\text{sim}}, \\[8pt]
        0.5 & \text{otherwise,}
    \end{cases}
\end{equation}
where $\mathcal{S}$ is the set of directly scored candidates, $\hat{r}_i$ their LLM scores, and $\text{Tan}(j,i)$ the Tanimoto coefficient between candidates $j$ and $i$. The fallback value of $0.5$ reflects genuine uncertainty rather than a positive or negative signal.

\paragraph{Surrogate update.}
All scores --- direct or imputed --- are used to update $s_{\text{llm}}(\mathbf{z})$  via $n_{\text{inner}}$ MSE gradient steps on stop-gradient latent copies, keeping surrogate updates decoupled from the main optimization graph. Between LLM calls, $s_{\text{llm}}(\mathbf{z})$ acts as a continuously queryable critic distilled from the episodic LLM feedback, providing a differentiable reward signal at every optimization step.

\section{Datasets}\label{app:datasets}
 To evaluate our method, we use cell lines contained in the NCI60 Growth Inhibition Data (accessed July 2022)~\cite{shoemaker2006nci60}, preprocessed at G2D-Diff~\cite{kim2025genotype}. The dataset binary genotype profile of a cancer cell line across $G=718$ clinically relevant genes. Optimization is performed per cell line across 15 cancer cell lines drawn from three held-out evaluation sets (EV1, EV2, EV3): EV1 contains data-rich cell lines seen during training, EV2 contains data-scarce cell lines with known conditions but unseen compound pairs, and EV3 is a fully zero-shot set of unseen cell lines(Table~\ref{tab:Evaluation Sets}).

\begin{table}
    \centering
       \caption{Details of the evaluation set}
    \label{tab:Evaluation Sets}
    \begin{tabular}{|c|c|}
    \hline
    Dataset                              & Cell Line\\ \hline
     \multirow{5}{*}{Evaluation Set 1}   & HCT116 \\
                                         & K562\\
                                         & LOXIMVI\\
                                         & MCF7\\
                                         & CCRFCEM \\
                                         \hline
     \multirow{5}{*}{Evaluation Set 2}   & EKVX\\
                                         & SKMEL28\\
                                         & SKOV3\\
                                         & NCIH226\\
                                         & OVCAR4 \\
                                         \hline
    \multirow{5}{*}{Evaluation Set 3}    & TK10 \\
                                         & OVCAR5 \\
                                         & HOP92 \\
                                         & SKMEL2 \\
                                         & HS578T\\
                                         \hline
    \end{tabular}
 
\end{table}

\section{Hyperparameter Search}
\label{app:hparams}

\subsection{Search Strategy}

We conducted a random hyperparameter search over six parameters governing the latent-space optimisation loop. Each candidate configuration was evaluated on three fixed batches sampled from the training cell pool (sensitive cell lines, AUC class 0 or 1, excluding validation and test sets). Using fixed batches ensures a fair, consistent comparison across all configurations. The search scoring metric is:
\begin{equation}
    \text{score} = -\overline{\text{AUC}} + 0.5 \cdot \overline{\text{QED}} - 0.5 \cdot \frac{\overline{\text{SAS}}}{10},
\end{equation}
where all values are ground-truth RDKit outputs (never surrogate estimates). This metric deliberately weights AUC most heavily, as sensitivity is the primary objective; QED and SAS contribute secondary regularisation.

\subsection{Search Space}

Table~\ref{tab:search_space} summarises the discrete values explored for each hyperparameter.

\begin{table}[h]
\centering
\caption{Hyperparameter search space. All combinations are explored via random sampling ($n_{\text{trials}}=20$, seed=42).}
\label{tab:search_space}
\resizebox{\textwidth}{!}{
\begin{tabular}{llp{7cm}}
\toprule
\textbf{Parameter} & \textbf{Values searched} & \textbf{Description} \\
\midrule
$\eta$ (learning rate) & \{1e-3, 3e-3, 5e-3, 1e-2, 2e-2\} & Adam learning rate for $\boldsymbol{\delta}$ \\
$\lambda$ ($\ell_2$ regularisation) & \{0.01, 0.03, 0.05, 0.10, 0.20\} & Penalty coefficient for $\|\boldsymbol{\delta}\|_2^2$ \\
$T$ (optimisation steps) & \{100, 150, 200, 300\} & Number of gradient ascent steps per batch \\
$S_{\text{ddim}}$ (diffusion steps) & \{30, 50\} & DDPM denoising steps during molecule generation \\
$\eta_s$ (surrogate learning rate) & \{1e-4, 5e-4, 1e-3, 3e-3\} & Adam learning rate for QED/SAS surrogates \\
$n_{\text{inner}}$ (surrogate inner steps) & \{2, 3, 5\} & Surrogate update steps per outer optimisation step \\
\bottomrule
\end{tabular}}
\end{table}

\subsection{Best Hyperparameters}

Table~\ref{tab:best_hparams} reports the selected configuration and its validation performance.

\begin{table}[h]
\centering
\caption{Best hyperparameter configuration selected by random search and corresponding validation metrics on the held-out training-pool batches.}
\label{tab:best_hparams}
\begin{tabular}{ll}
\toprule
\textbf{Hyperparameter} & \textbf{Selected value} \\
\midrule
Learning rate $\eta$ & 0.02 \\
$\ell_2$ regularisation $\lambda$ & 0.10 \\
Optimisation steps $T$ & 100 \\
Diffusion steps $S_{\text{ddim}}$ & 30 \\
Surrogate learning rate $\eta_s$ & 1e-4 \\
Surrogate inner steps $n_{\text{inner}}$ & 5 \\
Surrogate warmup $\tau_{\text{warm}}$ & 5 (fixed) \\
\midrule
\textbf{Reward weight} & \textbf{Value} \\
\midrule
$w_{\text{auc}}$ & 1.0 \\
$w_{\text{qed}}$ & 0.5 \\
$w_{\text{sas}}$ & 0.5 \\
$w_{\text{llm}}$ & 1.0 \\
\midrule
\textbf{Validation metric} & \textbf{Value} \\
\midrule
Search score & 0.004 \\
Mean predicted AUC & 0.117 \\
Mean QED & 0.565 \\
Mean SAS & 3.24 \\
SMILES validity & 1.00 \\
\bottomrule
\end{tabular}
\end{table}

\noindent The selected configuration achieves near-perfect SMILES validity ($= 1.0$), a low mean predicted AUC of 0.117 (well within the sensitive range), and a mean SAS of 3.24 (below the drug-like threshold of 4.5). The relatively short optimisation ($T=100$, $S_{\text{ddim}}=30$) indicates that meaningful latent improvements are achievable with a compact budget, consistent with the strong regularisation ($\lambda=0.10$) keeping $\boldsymbol{\delta}$ close to the DDPM prior.

\section{Baselines}~\label{app:baselines}

\subsection{G2D-Diff}

We used the weights pretrained in their work. The base model was trained with w=0 (no guidance scaling), but in the paper, they generated with w=7. This is a standard CFG practice — you train with dropout (prand=0.1 randomly zeros the condition) to enable unconditional predictions, then amplify guidance only at inference time. 

\subsection{PaccMannRL}
To ensure a fair comparison with our proposed method, we retrained the key components of the baselines. For PaccMannRL, the framework uses the same datasets and evaluation protocol adopted in this study. Specifically, we retained the pretrained molecular generative model (SVAE) trained on the ChEMBL dataset, as in the original PaccMannRL implementation, and re-trained the remaining components using the drug–response and omics data used for training our generative model. The PaccMann predictor was retrained using the cell line–centric drug response dataset, where drug sensitivity values were paired with cell line omics profiles and molecular SMILES representations. Consistent with the evaluation protocol, cell lines belonging to evaluation set 3 were excluded from predictor training to prevent information leakage. For the conditioning information, we constructed omics feature vectors by concatenating mutation, copy-number amplification, and copy-number deletion profiles for each cell line, which were then used to train the omics variational autoencoder (PVAE) responsible for encoding cellular context. The PVAE was trained on the same conditioning dataset used during generator training, excluding cell lines from evaluation set 2 in order to match the experimental design used for the proposed model. Subsequently, the PaccMannRL generator was trained using reinforcement learning with the retrained predictor as the reward function. As reported in prior work, the generator exhibited rapid mode collapse, and therefore RL optimization was limited to a single training epoch. 

\subsection{MolGen-GPT}
For MolGen-GPT, we do not retrain, as they already used the ChemBL dataset from which Guacamol is derived. 

\section{Ablation}\label{app:ablation}
We conducted ablation studies to assess the contribution of each component in our approach, presented in Table~\ref{tab:ablation}.

\begin{table}[!h]
\centering
\caption{Ablation study on the contribution of each optimization component. Validity, Uniqueness, Novelty, and Diversity are averaged across Eval \#1--3. To compute LLM Score, we choose the top 10 samples with a higher AUC predicted value for each method in each cell line in evaluation set.} 
\label{tab:ablation}
\resizebox{\textwidth}{!}{
\begin{tabular}{cccc|cccc|ccc|c}
\toprule
 \multicolumn{4}{c|}{Optimization Components} & \multicolumn{4}{c|}{Generation Quality} & \multicolumn{3}{c|}{Molecular Properties} & \multirow{3}{*}{LLM Score $\uparrow$} \\
\hline
 AUC & QED & SAS & LLM & Validity $\uparrow$ & Uniqueness $\uparrow$ & Novelty $\uparrow$ & Diversity $\uparrow$ & SAS $\downarrow$ & QED $\uparrow$ & LogP $\downarrow$ & \\
\midrule
— & — & — & — & \textbf{1.000} & 0.908 & 0.989 & 0.866 & 3.627 $\pm$ 1.038 & 0.346 $\pm$ 0.161 & 4.785 $\pm$ 1.948 & 0.353 $\pm$ 0.135 \\
 \checkmark & — & — & — & 0.684 & \textbf{0.999} & 0.999 & \textbf{0.880} & 4.145 $\pm$ 1.230 & 0.358 $\pm$ 0.175 & 4.708 $\pm$ 2.035 & 0.360 $\pm$ 0.157 \\
\checkmark & \checkmark & \checkmark & — & 0.999 & 0.994 & \textbf{1.000} & 0.870 & 3.059 $\pm$ 0.708 & 0.714 $\pm$ 0.178 & 3.438 $\pm$ 1.262 & 0.529 $\pm$ 0.151 \\
 \checkmark & \checkmark & \checkmark & \checkmark & \textbf{1.000} & 0.989 & 0.998 & 0.866 & \textbf{2.934} $\pm$ \textbf{0.672} & \textbf{0.747} $\pm$ \textbf{0.168} & \textbf{3.291} $\pm$ \textbf{1.198} & \textbf{0.569} $\pm$ \textbf{0.131} \\
\bottomrule
\end{tabular}
}

\end{table}
\section{Clinically Validated Drugs}\label{app:val_drugs}

We constructed an evaluation suite of 11 compound–cell-line pairs covering four FDA-approved targeted therapies with verified mechanism-of-action matches (imatinib~\cite{gambacorti2004development}/dasatinib~\cite{tokarski2006structure}, alpelisib~\cite{mayer2017phase}, vemurafenib~\cite{bollag2010clinical}), one mechanistic decoy (erlotinib~\cite{moyer1997induction}), one tool compound with poor drug-likeness (wortmannin~\cite{powis1994wortmannin}), one canonical PAINS scaffold (quinone~\cite{baell2010new}), and three structurally simple pharmacologically inert controls (aspirin~\cite{awtry2000aspirin}, metformin~\cite{bailey1996metformin}, caffeine~\cite{barone1996caffeine}, hexane~\cite{clough2014hexane}). 

\section{VAE Encoder Bias in Known-Binder Evaluation}
\label{app:encoder_bias}

\begin{figure}
    \centering
    \includegraphics[width=0.7\linewidth]{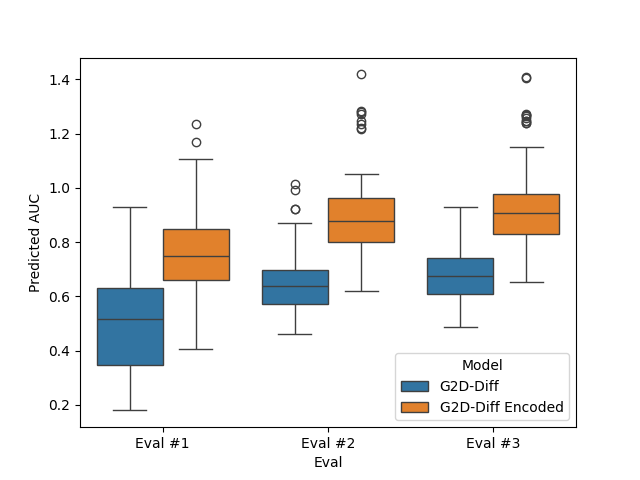}
    \caption{Distribution of predicted AUC values across evaluation sets for G2D-Diff and G2D-Diff (encoded).}
    \label{fig:encoded_results}
\end{figure}

The G2D-Diff VAE encoder is pretrained on a large corpus of drug-like molecules without supervision from cell-line drug response data, and its weights are kept frozen throughout both diffusion model training and latent space optimization.  As a result, the encoder maps molecules purely based on structural features,  with no awareness of the sensitivity-conditioned latent subspace that the diffusion model and the ensemble predictor operate within. This can be seen in Figure~\ref{fig:encoded_results}, where the molecules generated with G2D-Diff when encoded with VAE encoder give higher AUC values.

\end{document}